\documentclass[11pt,a4paper]{article}
\usepackage{times,latexsym}
\usepackage{url}
\usepackage[T1]{fontenc}
\usepackage{amssymb}
\usepackage[acceptedWithA]{tacl2021v1}

\usepackage{xspace,mfirstuc,tabulary}

\newif\iftaclinstructions
\taclinstructionsfalse 
\iftaclinstructions
\renewcommand{\confidential}{}
\renewcommand{\anonsubtext}{(No author info supplied here, for consistency with
TACL-submission anonymization requirements)}
\newcommand{\instr}
\fi

\iftaclpubformat 

\else

\fi

\usepackage{array}
\usepackage{booktabs}
\usepackage{multirow}
\usepackage[table]{xcolor}
\usepackage{threeparttable}
\usepackage{graphicx}
\usepackage{ulem} 
\usepackage{makecell}
\usepackage{adjustbox}
\usepackage{arydshln}
\usepackage{amsmath}
\usepackage{xcolor}
\usepackage{bm}
\definecolor{bestcolor}{RGB}{135, 175, 149} 
\definecolor{secondcolor}{RGB}{211, 234, 224} 
\definecolor{gaincolor}{RGB}{76,175,80}   
\definecolor{sealrow}{RGB}{212, 232, 225}

\newcommand{\MethodName}{AURORA}

\title{AURORA: Asymmetry and Update-Induced Rotation for Robust Hallucination Detection in Large Language Models}

\author{
Zishuai Zhang$^{1,2}$ \quad
Hainan Zhang\Thanks{Corresponding author.}  $^{1,2}$ \quad
Zhiming Zheng$^{1,2}$
\\
\\
$^1$School of Artificial Intelligence, Beihang University, China
\\
$^2$Beijing Advanced Innovation Center for Future Blockchain and Privacy Computing,
\\
Beihang University, China
\\
\texttt{zhangzishuai@buaa.edu.cn}\quad
\texttt{zhanghainan@buaa.edu.cn}\quad
\texttt{zhengzhiming0130@163.com}
}

\date{}

\begin{document}
\maketitle

\begin{abstract}
Large Language Models (LLMs) have demonstrated remarkable capabilities across a wide range of natural language processing tasks. However, their tendency to generate hallucinations, namely factually incorrect or unfaithful outputs, poses a critical obstacle to their deployment in high-stakes applications. Although recent hallucination detection methods have made encouraging progress, they typically rely on costly output-level consistency checks or static hidden-state probes that capture shallow dataset-specific patterns, leading to substantial degradation under cross-dataset evaluation.
In this work, we propose \MethodName{}, a novel hallucination detection framework that shifts the focus from static representations to the \textit{weight--gradient dynamics} of LLMs. Our key insight is that hallucinated and faithful answers induce qualitatively different gradient update patterns on the model's parameters. Specifically, hallucinated samples trigger asymmetric and structurally misaligned gradients, which can be captured through two complementary features: (1) the skewness of the cosine similarity distribution between weight matrices and their gradient update directions, and (2) the rotation ratio, which quantifies how much the gradient update reorients the singular-vector basis of weight matrices via SVD. \MethodName{} achieves strong hallucination detection performance across four model families and four benchmark datasets. Further analyses demonstrate that our method scales effectively across model sizes and transfers to out-of-domain tasks, including mathematical reasoning and vision-language scenarios.

\end{abstract}

\section{Introduction}\label{sec:Introduction}
Large Language Models (LLMs) have demonstrated unprecedented capabilities across a wide spectrum of natural language processing tasks, from question answering and summarization to code generation and scientific reasoning~\cite{grattafiori2024llama,yang2025qwen3technicalreport,gemmateam2025gemma3technicalreport}. However, their practical deployment in high-stakes domains---including healthcare, law, finance, and education---remains severely constrained by the persistent problem of \textit{hallucination}: the generation of outputs that are fluent and plausible-sounding, yet factually incorrect or unfaithful~\cite{li-etal-2023-halueval}. Reliable hallucination detection is therefore not only essential for improving the safety of LLM deployments, but also provides a principled tool for identifying abnormal generations, filtering unreliable data, and probing how models internally respond to faithful versus hallucinated content. Understanding such internal dynamics is particularly important for developing detection signals that generalize beyond dataset-specific surface patterns.

A diverse body of research has emerged to address this challenge, which can be broadly categorized along two axes. The first axis concerns the level of access to the model: \textit{black-box} methods operate solely on generated text, typically by assessing consistency among sampled responses or by validating a generated answer through query reconstruction and semantic comparison~\cite{manakul-etal-2023-selfcheckgpt,yehuda-etal-2024-interrogatellm}; \textit{white-box} methods leverage internal model signals such as hidden states, attention patterns, or output logits~\cite{zhang-etal-2025-icr,su-etal-2024-unsupervised}. The second axis concerns the nature of the detection signal: some methods~\cite{hu2025harphallucinationdetectionreasoning} extract features directly from forward-pass representations, while others rely on the model's uncertainty as reflected in token-level probabilities~\cite{farquhar2024detecting,fadeeva-etal-2024-fact}.

\begin{figure}[!ht]
  \centering
  \includegraphics[width=\linewidth]{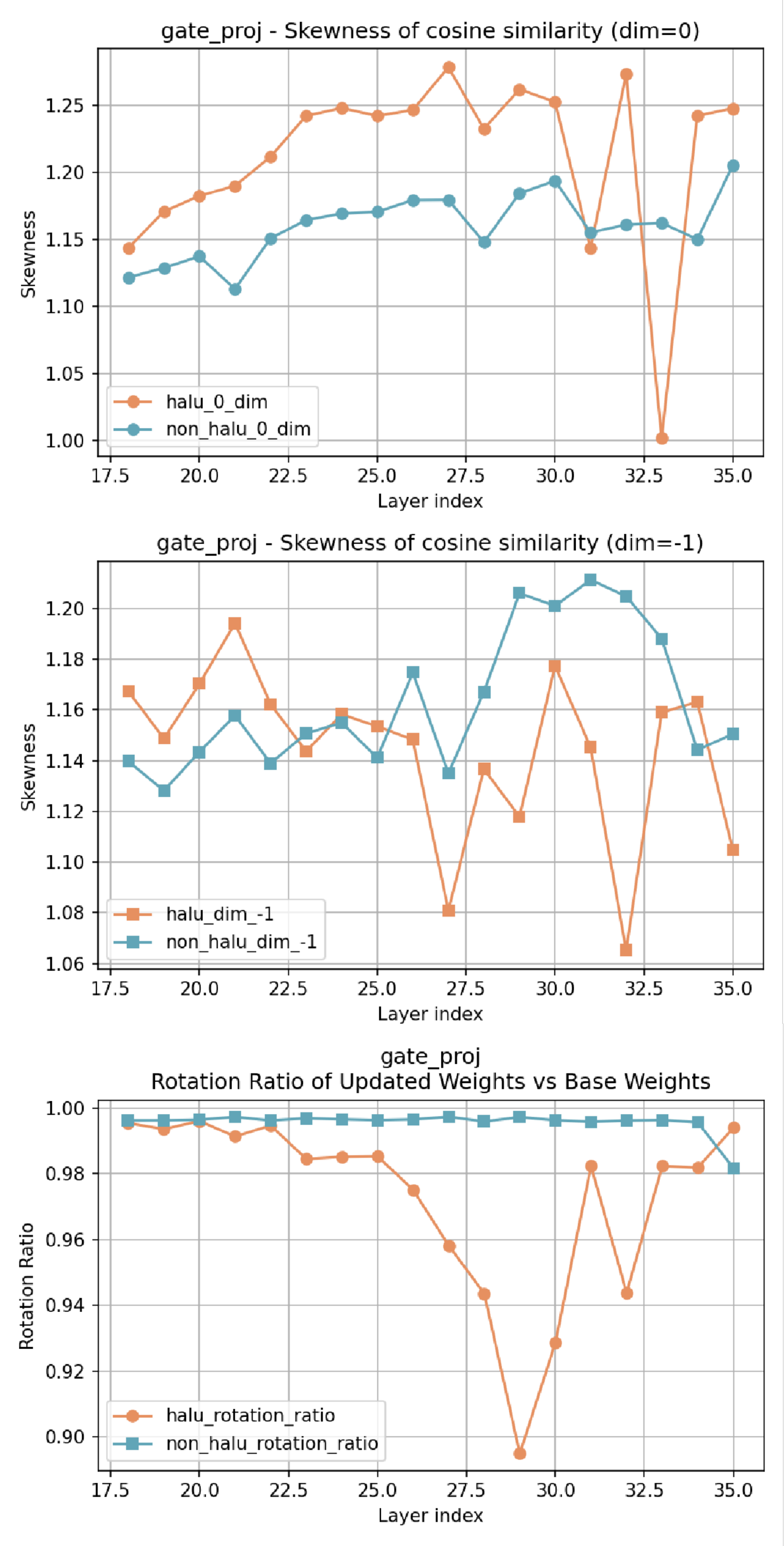}
  \caption{Hallucination vs. correct: Faithful answers induce relatively symmetric weight--update alignment with modest rotation from the original singular vector basis. Hallucinated answers produce skewed alignments and substantial basis rotation, reflecting the model's internal tension when fabricating information.}
  \label{fig:motivation}
\end{figure}

Despite the steady progress, most existing white-box methods share a fundamental limitation: they operate exclusively on \textit{static representations} extracted during the forward pass---hidden states, attention maps, or output embeddings. While such representations encode what the model ``knows'' at inference time, they do not capture how the model would \textit{change} if it were updated to better fit a given sample. We argue that this \textit{gradient-level} information---the direction of parameter updates induced by a specific input---carries a complementary and more generalizable signal for hallucination detection. Intuitively, when an LLM produces a faithful answer grounded in its parametric knowledge, the gradient update should be relatively aligned with the existing weight matrices. In contrast, when the model fabricates information, the resulting gradient signal should exhibit pronounced bias and structural misalignment, as the model attempts to reconcile an internally inconsistent generation with its learned parameterization. To verify this hypothesis, we first finetune Qwen3-8B on hallucinated and correct samples, respectively. We then compute the layer-wise skewness and the rotation ratio between the finetuned model and the base model. The results are shown in Figure~\ref{fig:motivation}. It can be observed that finetuning on hallucinated samples leads to noticeably larger weight rotations than finetuning on correct samples, particularly in the deeper transformer layers. In addition, the cosine-similarity distribution exhibits consistently higher skewness along the output dimension, suggesting that hallucination learning updates are concentrated on fewer dominant directions rather than being evenly distributed. These observations indicate that hallucinated samples induce more anisotropic parameter updates, motivating us to exploit skewness and weight rotation as potential indicators for hallucination detection.

Building on this insight, we propose \textbf{\MethodName{}}, a novel white-box hallucination detection framework that, to the best of our knowledge, is the first to exploit \textit{weight--update dynamics} for this task. Rather than inspecting hidden states, \MethodName{} extracts features from the relationship between a frozen LLM's weight matrices and the gradient update directions induced by individual training samples. Concretely, we equip the LLM with lightweight LoRA adapters~\cite{hu2022lora} and, for each query--answer pair, perform a forward--backward pass to obtain gradients. From these gradients, we compute two complementary types of features for each target layer:

\begin{enumerate}
    \item \textbf{Skewness of Cosine Similarity Distribution:} We compute the row-wise and column-wise cosine similarity between the current weight matrix and its gradient-updated counterpart, then measure the skewness of the resulting similarity distribution. A highly skewed distribution indicates that the gradient update asymmetrically affects different output or input dimensions, which often is a signature of hallucinated content.
    \item \textbf{Rotation Ratio via SVD:} We decompose each target weight matrix via SVD, project the gradient-direction weight onto the original singular subspaces, and quantify the off-diagonal energy, namely, rotation ratio. A high rotation ratio indicates that the gradient update substantially reorients the weight matrix away from its learned singular vector basis, reflecting the model's internal inconsistency when processing hallucinated answers.
\end{enumerate}

These features are standardized via per-layer, per-module z-score normalization and then fed into a MLP classifier. The full pipeline operates in two stages: (i) feature collection, where the frozen LLM with LoRA adapters processes training samples to extract gradient-level features, and (ii) probe training, where the MLP classifier is trained on these pre-extracted features with binary cross-entropy loss.

We conduct comprehensive experiments across four model families and different scales and four standard hallucination benchmarks (HaluEval, HotpotQA, TriviaQA, and SQuAD). Our main findings are:

\begin{itemize}
      \item \textbf{Novel features from weight update dynamics.}
      We discover that hallucinated and faithful samples induce qualitatively different
      parameter update patterns: the skewness of cosine similarity between weights and their gradient updates, and a rotation ratio via  SVD. Unlike prior work that probes
      static hidden states, our features characterize the \textit{dynamic interaction}
      between model knowledge and outer claims.

      \item \textbf{Superior cross-dataset generalization.}
      A MLP classifier trained on these features achieves competitive in-domain
      accuracy and substantially outperforms existing white-box methods under cross-dataset
      transfer. This pronounced
      transferability suggests that gradient-level features capture a more fundamental,
      model-intrinsic signature of hallucination rather than dataset-specific correlations.

      \item \textbf{Extensive validation across scales, architectures, and modalities.}
      We verify the effectiveness of \MethodName{} across model sizes spanning 0.8B to 70B
      parameters, and out-of-domain tasks
      including mathematical reasoning and vision-language scenarios. Ablation studies
      further reveal the complementary roles of individual features and identify the most
      informative transformer components for detection.
\end{itemize}

The rest of the paper proceeds as follows. Section~\ref{sec:Methodology} introduces our weight--update feature extraction framework and the downstream probe. Section~\ref{sec:Experiments} presents in-domain and cross-dataset evaluations, ablation analyses of individual features, layers and modules, together with cross-model scaling and out-of-domain transfer experiments. Section~\ref{sec:Conclusion} concludes.

\section{Related Work}\label{sec:Relatedwork}
Hallucination detection aims to determine whether an LLM-generated answer is faithful to the given query, context, or underlying facts. Existing methods can be broadly organized according to where the detection signal is obtained: output-level behavior, forward-pass internal representations, or spectral structures of model states. 

\paragraph{Output-level consistency and uncertainty methods.}
A common line of work treats the LLM as a black box and detects hallucinations from generated text alone. SelfCheckGPT~\cite{manakul-etal-2023-selfcheckgpt} is based on the intuition that factual answers should remain consistent across multiple stochastic generations, whereas hallucinated answers tend to produce divergent or contradictory samples. InterrogateLLM~\cite{yehuda-etal-2024-interrogatellm} instead validates a generated answer by asking the model to reconstruct the original query from that answer and then measuring semantic consistency between the reconstructed and original queries. Related uncertainty-based methods estimate hallucination risk from token-level probabilities~\cite{fadeeva-etal-2024-fact}, semantic entropy over sampled answers~\cite{farquhar2024detecting}, or embedding-space variability across sampled generations~\cite{chen2024inside}. These methods are attractive because they require little or no access to model internals. However, many of them require multiple generations, external entailment or embedding models, or carefully designed prompts. Moreover, because their signals are derived from output variability, they may conflate hallucination with benign diversity among plausible answers.

\paragraph{White-box methods based on forward-pass representations.}
Another major direction exploits internal model signals, including hidden states, contextualized embeddings, attention patterns, and logits. MIND~\cite{su-etal-2024-unsupervised} uses internal states to construct an unsupervised detector for real-time hallucination detection. ICR Probe~\cite{zhang-etal-2025-icr} tracks the evolution of hidden states across layers and measures the contribution of different modules to residual-stream updates. Compared with black-box methods, these approaches are often more sample-efficient and can operate with a single forward pass. However, because these methods rely on forward-pass representations under fixed model parameters, they may capture dataset-specific surface patterns rather than general hallucination-related signals. As a result, they often achieve strong in-domain performance but degrade under cross-dataset evaluation, indicating limited ability to identify hallucinations beyond the training distribution.

\paragraph{Spectral and subspace-based detection.}
Spectral methods provide another way to characterize hallucination-related structures in LLMs. HARP~\cite{hu2025harphallucinationdetectionreasoning} decomposes the unembedding matrix with SVD and projects hidden states onto a reasoning subspace for hallucination detection. EigenScore~\cite{chen2024inside} analyze the covariance or effective rank of embeddings obtained from multiple sampled generations.  Although these methods demonstrate the usefulness of spectral signals, they mainly decompose forward-pass activations, sampled-response embeddings, attention structures, or the unembedding matrix. In contrast, \MethodName{} applies SVD to weight matrices and measures how sample-induced gradient updates reorient their singular-vector basis. Therefore, our use of SVD is not merely a spectral summary of representations, but a way to quantify structural changes in the model's parameter geometry.

\MethodName{} differs from prior detectors by exploiting weight-update dynamics rather than output consistency or static forward-pass representations. It characterizes the interaction between weights and sample-induced gradient updates through two features: skewness, which measures asymmetric weight-gradient alignment, and rotation ratio, which quantifies the reorientation of the original singular-vector basis. By capturing how the model would change in response to each sample, \MethodName{} avoids repeated sampling, is less dependent on dataset-specific hidden-state patterns, and provides a more transferable signal for hallucination detection and abnormal data filtering.

\section{Methodology}\label{sec:Methodology}
\subsection{Problem Formulation}
Let a dataset $\mathcal{D} = \{(\mathbf{q}_i, \mathbf{a}_i, y_i)\}_{i=1}^{N}$ consist of user queries $\mathbf{q}_i$ and LLM-generated answers $\mathbf{a}_i$, where each pair is labeled $y_i \in \{0, 1\}$ indicating whether $\mathbf{a}_i$ is faithful ($y_i = 0$) or hallucinated ($y_i = 1$) with respect to $\mathbf{q}_i$. Our goal is to learn a binary classifier
\begin{equation}
f_\theta(\mathbf{q}, \mathbf{a}) \;\to\; \{0, 1\}
\end{equation}
that predicts whether a given answer is hallucinated. The classifier $f_\theta$ does not operate on raw text directly. Instead, it consumes a compact feature vector $\mathbf{z} \in \mathbb{R}^F$ extracted from the internal weight--gradient dynamics of a trainable LLM. Formally,
\begin{equation}
f_\theta(\mathbf{q}, \mathbf{a}) = f_\theta\bigl(\Phi(\mathbf{q}, \mathbf{a}; \mathcal{M})\bigr),
\end{equation}
where $\mathcal{M}$ denotes the LLM adapted via LoRA~\cite{hu2022lora}, which is employed to keep the training lightweight while efficiently obtaining the model update direction. $\Phi(\cdot)$ denotes the feature extractor (detailed in \ref{sec:feature_extraction}), and $f_\theta$ is the lightweight MLP trained with binary cross-entropy loss.

This formulation supports two practical deployment scenarios. \textbf{(1) Post-hoc hallucination detection:} given a deployed QA system, after the LLM produces an answer $\mathbf{a}$ to a user query $\mathbf{q}$, $f_\theta$ can flag the answer as potentially hallucinated before it reaches the end user. \textbf{(2) Training data filtering:} when constructing supervised fine-tuning datasets from LLM-generated outputs, one can score each $(\mathbf{q}, \mathbf{a})$ pair using $f_\theta$ and discard samples whose predicted hallucination probability exceeds a threshold, thereby improving training data quality without manual verification.

\subsection{\MethodName{}}\label{sec:feature_extraction}
We extract features from the LoRA-adapted layers of the LLM. For each target module (e.g., \texttt{v\_proj}, \texttt{gate\_proj}), let $\mathbf{W}_0 \in \mathbb{R}^{d_{\text{out}} \times d_{\text{in}}}$ denote the frozen pre-trained weight matrix. LoRA introduces two low-rank matrices $\mathbf{B} \in \mathbb{R}^{d_{\text{out}} \times r}$ and $\mathbf{A} \in \mathbb{R}^{r \times d_{\text{in}}}$, where $\mathbf{B}$ is initialized to zero and $\mathbf{A}$ is randomly initialized. The effective weight matrix is:
\begin{equation}
\mathbf{W} = \mathbf{W}_0 + \frac{\alpha}{r}\mathbf{B}\mathbf{A}.
\end{equation}
Given an input pair $(\mathbf{q}, \mathbf{a})$, we perform a forward pass through the model to compute the language modeling loss $\mathcal{L}(\mathbf{a} \mid \mathbf{q}; \mathbf{W})$. We then perform a backward pass, which yields gradients $\nabla_{\mathbf{B}}\mathcal{L}$ and $\nabla_{\mathbf{A}}\mathcal{L}$ with respect to the LoRA parameters. Since $\mathbf{W}_0$ is frozen, the effective weight change induced by one gradient step is propagated solely through the low-rank adapters. We approximate this change via a first-order expansion:

\begin{equation}
\delta\mathbf{W} = (\nabla_{\mathbf{B}}\mathcal{L})\,\mathbf{A} \;+\; \mathbf{B}\,(\nabla_{\mathbf{A}}\mathcal{L}),
\label{eq:deltaW}
\end{equation}

where we omit the second-order term $(\nabla_{\mathbf{B}}\mathcal{L})(\nabla_{\mathbf{A}}\mathcal{L})$, which is negligible relative to the first-order contributions. Equation~\eqref{eq:deltaW} captures how the effective weight $\mathbf{W}$ would shift under one gradient step, projected through the low-rank LoRA subspace. We then define the gradient-direction weight as:
\begin{equation}
\mathbf{G} = \mathbf{W} - \gamma \cdot \delta\mathbf{W},
\label{eq:G_def}
\end{equation}
representing the hypothetical weight matrix after a single gradient update, where $\gamma$ is the strength factor. All subsequent features are derived from the pair $(\mathbf{W}, \mathbf{G})$.

\paragraph{Skewness of Cosine Similarity Distribution}

For each module, we compute the row-wise cosine similarity between $\mathbf{W}$ and $\mathbf{G}$:

\begin{equation}
\cos\theta_i^{(\text{row})} = \frac{\mathbf{W}_{i,:} \cdot \mathbf{G}_{i,:}}{\|\mathbf{W}_{i,:}\| \; \|\mathbf{G}_{i,:}\|}, \quad i = 1, \dots, d_{\text{out}}.
\label{eq:cos_row}
\end{equation}

This yields a distribution of $d_{\text{out}}$ cosine similarity values, each quantifying how much the $i$-th output direction rotates under the gradient update. We compute the skewness of this distribution:

\begin{equation}
\gamma^{(\text{row})} = \frac{1}{d_{\text{out}}}\sum_{i=1}^{d_{\text{out}}} \left(\frac{\cos\theta_i^{(\text{row})} - \mu_{\text{row}}}{\sigma_{\text{row}}}\right)^3,
\label{eq:skew_row}
\end{equation}

where $\mu_{\text{row}}$ and $\sigma_{\text{row}}$ are the empirical mean and standard deviation of $\{\cos\theta_i^{(\text{row})}\}$. Analogously, we compute column-wise cosine similarities along the input dimension $d_{\text{in}}$ (replacing $\mathbf{W}_{i,:}$ with $\mathbf{W}_{:,j}$ in Eq.~\eqref{eq:cos_row}), producing a column-wise skewness $\gamma^{(\text{col})}$.

These two skewness measures capture complementary asymmetry patterns:
\begin{itemize}
    \item \textbf{Row-wise skewness} ($\gamma^{(\text{row})}$, referred to as \texttt{skew\_dim\_minus1} in our code): a highly positive skew means a small fraction of output dimensions experience disproportionately large directional shifts, while most remain nearly unchanged. A negative skew indicates that the directional shifts are more uniformly spread across output dimensions, with a long tail of dimensions shifting in the opposite direction.
    \item \textbf{Column-wise skewness} ($\gamma^{(\text{col})}$, referred to as \texttt{skew\_dim\_0}): analogously measures the asymmetry across input channels, reflecting how selectively different input features contribute to the gradient-induced rotation.
\end{itemize}

After extraction, each skewness value is standardized via per-layer, per-module z-score normalization using pre-computed mean and standard deviation statistics (see Appendix~\ref{sec:Appendix_Main_results}).

\paragraph{Rotation Ratio via SVD}
While skewness captures per-vector asymmetry, the rotation ratio measures the \textit{global structural rotation} of the weight matrix induced by the gradient update. Given the effective weight $\mathbf{W} \in \mathbb{R}^{d_{\text{out}} \times d_{\text{in}}}$, we perform rank-$r$ truncated SVD:

\begin{equation}
\mathbf{W} \approx \mathbf{U}_r \mathbf{S}_r \mathbf{V}_r^\top,
\label{eq:svd}
\end{equation}

where $\mathbf{U}_r \in \mathbb{R}^{d_{\text{out}} \times r}$ and $\mathbf{V}_r \in \mathbb{R}^{d_{\text{in}} \times r}$ have orthonormal columns spanning the top-$r$ left and right singular subspaces, and $\mathbf{S}_r \in \mathbb{R}^{r \times r}$ is diagonal. We then project the gradient-direction weight $\mathbf{G}$ (Eq.~\eqref{eq:G_def}) onto these original singular subspaces:

\begin{equation}
\tilde{\mathbf{S}} = \mathbf{U}_r^\top \mathbf{G} \mathbf{V}_r.
\label{eq:S_tilde}
\end{equation}

If the gradient update merely rescales the existing singular directions without rotating them, $\tilde{\mathbf{S}}$ remains diagonal. Conversely, if the update introduces new directions not aligned with $\mathbf{U}_r$ or $\mathbf{V}_r$, the off-diagonal entries of $\tilde{\mathbf{S}}$ become non-zero. We quantify this via the rotation ratio:

\begin{equation}
\rho = \lambda \cdot \frac{\bigl\|\tilde{\mathbf{S}} - \operatorname{diag}(\tilde{\mathbf{S}})\bigr\|_F}{\|\tilde{\mathbf{S}}\|_F + \epsilon},
\label{eq:rotation_ratio}
\end{equation}

where $\lambda = 500$ is a scaling constant (chosen to bring the raw ratio, typically in $[0, 0.002]$, to a numerically convenient range), $\|\cdot\|_F$ denotes the Frobenius norm, and $\epsilon = 10^{-8}$ prevents division by zero. A higher $\rho$ indicates that the gradient update substantially rotates the weight matrix away from its original singular vector basis, reorienting the model's internal representations. The rotation ratio is also standardized via per-layer, per-module z-score normalization.

The weights of the LLMs are high-dimensional, leading to prohibitive computational cost when performing SVD for downstream analysis. To mitigate this issue, we adopt truncated SVD, which approximates the original feature matrix by retaining only the top r singular components. This not only reduces the computational overhead but also highlights the most discriminative directions.
\section{Experiments}\label{sec:Experiments}

\subsection{Experiment Setup}

\paragraph{Datasets}
\begin{itemize}
    \item \textbf{HaluEval}~\cite{li-etal-2023-halueval}: a large-scale hallucination evaluation benchmark containing both faithful and hallucinated responses.
    \item \textbf{HotpotQA}~\cite{yang2018hotpotqa}: a multi-hop QA dataset that requires reasoning over multiple supporting facts.
    \item \textbf{TriviaQA}~\cite{JoshiTriviaQA2017}: an open-domain QA benchmark with diverse fact-seeking questions.
    \item \textbf{SQuAD}~\cite{rajpurkar-etal-2018-know}: a reading comprehension benchmark for evaluating answer faithfulness to given passages.  
\end{itemize}

\paragraph{Metrics}
\begin{itemize}
      \item \textbf{Balanced Accuracy (B-ACC).}
      We use balanced accuracy
      as the primary metric to account for potential class imbalance:
      $\text{B-ACC} = \frac{1}{2}\bigl(\frac{\text{TP}}{\text{TP}+\text{FN}}
      + \frac{\text{TN}}{\text{TN}+\text{FP}}\bigr)$.
      The decision threshold is fixed at 0.5 for all \MethodName{} experiments. Unless otherwise specified by the original paper, we report B-ACC at the
      optimal threshold for baselines.

      \item \textbf{Generalization Accuracy (Gen-ACC).}
      For cross-dataset evaluation, we compute the mean value of B-ACC across
      the four evaluation datasets using the same checkpoints, weighting each dataset equally:
      $\text{Gen-ACC} = \frac{1}{4}\sum_{d=1}^{4} \text{B-ACC}_d$.

\end{itemize}

\paragraph{Baselines}
\begin{itemize}
    \item \textbf{ICR Probe}~\cite{zhang-etal-2025-icr}: a white-box detector that tracks the cross-layer evolution of hidden states. It computes the Information contribution to residual stream (ICR) score to quantify how different modules contribute to hidden-state updates, and trains a lightweight probe for hallucination classification.
    
    \item \textbf{HARP}~\cite{hu2025harphallucinationdetectionreasoning}: a reasoning-subspace based detector. It applies SVD to the unembedding matrix to separate semantic and reasoning subspaces, projects hidden states onto the reasoning subspace, and uses the resulting projections as hallucination detection features.
    
    \item \textbf{InterrogateLLM}~\cite{yehuda-etal-2024-interrogatellm}: a zero-resource black-box method that validates a generated answer through backward query reconstruction. It asks the LLM to reconstruct the original query from the generated answer and measures the semantic consistency between the reconstructed and original queries.

\end{itemize}

\paragraph{Parameter Settings}
We conduct experiments on four widely adopted LLMs, including Qwen3-8B-Instruct\cite{yang2025qwen3technicalreport}, LLaMA3-8B-Instruct\cite{grattafiori2024llama}, Ministral3-8B-Base-2512\cite{liu2026ministral3}, and Gemma-3-12B-Instruct\cite{gemmateam2025gemma3technicalreport}. These models cover both base and instruction-tuned variants, while also spanning different architectural families.

We train \MethodName{} and all baseline LLMs in BF16, using LoRA with rank 16 and $\alpha$ = 32. All models are optimized with a learning rate of 1e-5 and trained with early stopping to ensure that each baseline achieves its best performance. For \MethodName{}, the classifier is trained with a learning rate of 3e-5, using 16 layers with a hidden dimension of 4096 and an SVD rank of 128. Other parameters are in \ref{sec:Appendix_Main_results}.

\subsection{Cross-dataset Results}
\begin{table*}[!th]
  \centering
  \caption{Performance comparison across four datasets.}
  \label{tab:generalization_appendix}
  \begin{adjustbox}{max width=\linewidth}
    \begin{tabular}{lccccccc}
      \toprule
      \multirow{2}{*}{Method}&\multirow{2}{*}{Train dataset}&\multicolumn{4}{c}{Test dataset}&\multirow{2}{*}{Avg. \textcolor{green!60!black}{\(\large\bm{\uparrow}\)}}&\multirow{2}{*}{Std. \textcolor{green!60!black}{\(\large\bm{\downarrow}\)}}\\
      \cline{3-6}
       &  & HaluEval & HotpotQA & TriviaQA & SQuAD &  &  \\
      \midrule
      \rowcolor{sealrow} \multicolumn{8}{l}{\textbf{\textit{Llama-3-8B-it}}} \\
      HARP &\multirow{2}{*}{HaluEval} & \textbf{0.987} & 0.569 & 0.5635 & 0.5615 &0.6703&0.0335\\
      Ours  & & 0.9615 & \textbf{0.7610} & \textbf{0.8340} & \textbf{0.6540} & \textbf{0.8026} & 0.0125 \\
      \hdashline
      HARP& \multirow{2}{*}{HotpotQA} & \textbf{0.647} & \textbf{0.9670} & \textbf{0.9720} & 0.609 & \textbf{0.7988} &0.0293\\
      Ours &  & 0.5625 & 0.9620 & 0.9670 & \textbf{0.6375} & 0.7823 & 0.0339 \\
      \hdashline
      HARP& \multirow{2}{*}{TriviaQA} & \textbf{0.72}& 0.825 & 0.9095&0.5935 &0.7620&0.0140\\
      Ours &  & 0.5850 & \textbf{0.9450} & \textbf{0.9795} & \textbf{0.6480} & \textbf{0.7894} & 0.0305 \\
      \hdashline
      ICR & \multirow{3}{*}{SQuAD} & 0.5665 & 0.7185 & 0.7180 & 0.7575 & 0.6901 & 0.0054 \\
      HARP & & \textbf{0.779} & 0.6765 & 0.7330 & \textbf{0.8540} & 0.7606 & 0.0042 \\
      Ours & & 0.7285 & \textbf{0.7415} & \textbf{0.7915} & 0.8405 & \textbf{0.7755} & 0.0020 \\
      \hline
      
      \rowcolor{sealrow} \multicolumn{8}{l}{\textbf{\textit{Qwen-3-8B-it}}} \\
      HARP&\multirow{2}{*}{HaluEval} & \textbf{0.9905} &0.8275&0.845&0.5935&0.8141&0.0202\\
      Ours&  & 0.9505 & \textbf{0.8305} & \textbf{0.8910} & \textbf{0.6815} & \textbf{0.8384} & 0.0100 \\
      \hdashline
      HARP& \multirow{2}{*}{HotpotQA} &\textbf{0.6865}&\textbf{0.9805} & \textbf{0.9885} & 0.621 & 0.8191 & 0.0279 \\
      Ours& & 0.6460 & 0.9660 & 0.9680 & \textbf{0.7080} & \textbf{0.8220} & 0.0215 \\
      \hdashline
      HARP& \multirow{2}{*}{TriviaQA}&0.6755&\textbf{0.9625}&\textbf{0.9895}&0.6375&0.8163&0.0258\\
      Ours&  & \textbf{0.7680} & 0.9390 & 0.9780 & \textbf{0.6960} & \textbf{0.8453} & 0.0137 \\
      \hdashline
      ICR & \multirow{3}{*}{SQuAD} & 0.7785 & 0.5840 & 0.6075 & 0.6600 & 0.6575 & 0.0056\\
      HARP & &\textbf{0.9045} & \textbf{0.8825} & \textbf{0.9445} & \textbf{0.872} & \textbf{0.9009} & 0.0008 \\
      Ours & & 0.8050 & 0.8580 & 0.9100 & 0.8335 & 0.8516 & 0.0015 \\
      \hline
      \rowcolor{sealrow} \multicolumn{8}{l}{\textbf{\textit{Ministral3-8B-Base}}} \\
      ICR & \multirow{3}{*}{HaluEval}&0.9590 & 0.6170 & 0.5945 & 0.5150 & 0.6714 & 0.0290 \\
      HARP & & \textbf{0.985} & 0.6835 & 0.6621 & 0.535 & 0.7164 & 0.0273 \\
      Ours & & 0.9695 & \textbf{0.8500} & \textbf{0.9205} & \textbf{0.6975} & \textbf{0.8594} & 0.0105 \\
      \hdashline
      HARP& \multirow{2}{*}{HotpotQA} & \textbf{0.8045} &0.919&0.9185&0.64&0.8205&0.0130\\
      Ours&  & 0.6730 & \textbf{0.9775} & \textbf{0.9760} & \textbf{0.6955} & \textbf{0.8305} & 0.0215 \\
      \hdashline
      HARP& \multirow{2}{*}{TriviaQA} & \textbf{0.8125} &0.8945&0.9276&0.617&0.8129&0.0145 \\
      Ours&  & 0.6680 & \textbf{0.9480} & \textbf{0.9830} & \textbf{0.7285} & \textbf{0.8319} & 0.0185 \\
      \hdashline
      HARP& \multirow{2}{*}{SQuAD} & \textbf{0.8765} &0.7085&0.8546&0.8975&0.8343 & 0.0055\\
      Ours&  & 0.7435 & \textbf{0.8495} & \textbf{0.9070} & \textbf{0.8395} & \textbf{0.8349} & 0.0034 \\
      \hdashline
      \rowcolor{sealrow} \multicolumn{8}{l}{\textbf{\textit{Gemma3-12B-it}}} \\
      HARP& \multirow{2}{*}{HaluEval} &\textbf{0.987}&0.8755&0.8785&0.6645&0.8514&0.0137\\
      Ours&  & 0.9780 & \textbf{0.8910} & \textbf{0.9250} & \textbf{0.8135} & \textbf{0.9019} & 0.0036 \\
      \hdashline
      HARP& \multirow{2}{*}{HotpotQA} &0.6335&0.9565&0.9565&0.7735&0.8300&0.0185\\
      Ours&  & \textbf{0.7500} & \textbf{0.9795} & \textbf{0.9835} & \textbf{0.8385} & \textbf{0.8879} & 0.0097 \\
      \hdashline
      ICR& \multirow{3}{*}{HotpotQA} & 0.6215 & \textbf{0.9595} & 0.9710 & 0.5125 & 0.7661 & 0.0412 \\
      HARP& & 0.6975 & 0.953 & 0.98 & 0.7885 & 0.8548 & 0.0136 \\
      Ours& & \textbf{0.8000} & 0.9480 & \textbf{0.9895} & \textbf{0.8200} & \textbf{0.8894} & 0.0066 \\
      \hdashline
      HARP& \multirow{2}{*}{SQuAD} &0.831&0.875&0.9205& \textbf{0.917} &0.8859&0.0013\\
      Ours& & \textbf{0.8300} & \textbf{0.8855} & \textbf{0.9325} & 0.9100 & \textbf{0.8895} & 0.0015 \\
      \bottomrule
    \end{tabular}
  \end{adjustbox}
\end{table*}

The generalization results is shown in \ref{tab:generalization_appendix}.

\subsection{In-domain Results}
The in-domain results are shown in \ref{In-domain Results}.
\begin{table*}[htbp]
  \caption{In-domain Results}
  \label{In-domain Results}
  \centering
  \begin{adjustbox}{max width=\linewidth}
    \footnotesize
    \setlength{\tabcolsep}{5pt}
    \renewcommand{\arraystretch}{1.15}
    \begin{tabular}{lcccccc}
      \toprule
      \multirow{2}{*}{\textbf{Model}}
      & \multicolumn{1}{c}{\textbf{HaluEval}} & \multicolumn{1}{c}{\textbf{HotpotQA}} & \multicolumn{1}{c}{\textbf{TriviaQA}} & \multicolumn{1}{c}{\textbf{SQuAD}} & \multicolumn{2}{c}{\textbf{Avg}} \\
      \cmidrule(lr){2-2} \cmidrule(lr){3-3} \cmidrule(lr){4-4} \cmidrule(lr){5-5} \cmidrule(lr){6-6}
      & \textbf{\% Acc.} &\textbf{\% Acc.} & \textbf{\% Acc.} &\textbf{\% Acc.}  & \textbf{\% Acc.}\\
      \hdashline
    
      \rowcolor{sealrow} \textbf{\textit{Llama-3-8B-it}}
      & & & & & \\
      \quad + ICR probe
      & 0.9625 & 0.9355 & 0.9375 & 0.7880 & 0.9059 \\
      \quad + InterrogateLLM
      & 0.5050 & 0.6050 & 0.573 & 0.696 & 0.5948 \\
      \quad + HARP
      & 0.986 & 0.9675 & 0.982 &0.8745&0.9525 \\
      \quad + \MethodName{}
      & 0.9675
      & 0.9640
      & 0.9825
      & 0.8520
      & 0.9415 \\
      \midrule

      \rowcolor{sealrow} \textbf{\textit{Gemma3-12B-it}}
      & & & & &\\
      \quad + InterrogateLLM
      &  0.5705 & 0.595  & 0.6905  & 0.672 & 0.6320 \\
      \quad + ICR probe
      & 0.9775 & 0.9710 & 0.9765 &0.7325& 0.9144\\
      \quad + HARP
      & 0.989 & 0.983 & 0.9865 & 0.9195 & 0.9695 \\
      \quad + \MethodName{}
      & 0.9800
      & 0.9825
      & 0.9910
      & 0.9220
      & 0.9689 \\
      \midrule

      \rowcolor{sealrow} \textbf{\textit{Ministral3-8B-Base}}
      & & & & &\\
      \quad + InterrogateLLM
      &  0.5  & 0.677 &  0.841 & 0.642 & 0.6650 \\
      \quad + ICR probe
      & 0.9655 & 0.9280 & 0.9340 & 0.7675 & 0.8988 \\
      \quad + HARP
      & 0.984 & 0.9795 & 0.9905 & 0.8925 & 0.9616\\
      \quad + \MethodName{}
      & 0.9730
      & 0.9820
      & 0.9845
      & 0.8505
      & 0.9475 \\
      \midrule

      \rowcolor{sealrow} \textbf{\textit{Qwen3-8B-it}}
      & & & & & \\
      \quad + InterrogateLLM
      & 0.5 & 0.683 & 0.7645 & 0.728 & 0.6689 \\
      \quad + ICR probe
      & 0.9680 & 0.9475& 0.9500 & 0.6815 & 0.8868 \\
      \quad + HARP
      & 0.991 & 0.9805 &0.99 &0.8705 &0.9580 \\
      \quad + \MethodName{}
      & 0.9580
      & 0.9705
      & 0.9790
      & 0.8520
      & 0.9399 \\
      \bottomrule
    \end{tabular}
  \end{adjustbox}
\end{table*}

\subsection{Analysis}
\subsubsection{Ablation Study}
\paragraph{Ablation on Individual Features}To assess the independent contribution of each feature, we train separate models where only one feature group is provided as input. Specifically, we fine-tune Gemma3-12B-it on the HaluEval training set for 5 epochs with a learning rate of 3e-5. We report the highest Acc. on the HaluEval test set, and the highest Generalization Acc. across the four test sets. Table \ref{tab:feature_performance} summarizes the results. The results show that jointly leveraging all features achieves the best overall performance, with an in-domain accuracy of 0.9810 and a generalization accuracy of 0.8914, confirming the benefits of feature complementarity. Among the individual feature groups, skewness\_0 yields the strongest standalone performance, which indicates that it captures the most direct hallucination-related signal. Removing skewness\_0 causes the largest drop in in-domain accuracy (0.9660), further validating its central role. In contrast, excluding rotation\_ratio leaves the HaluEval accuracy unchanged at 0.9810 but reduces generalization to 0.8796, revealing that rotation\_ratio primarily enhances out-of-distribution robustness rather than in-domain separation. Removing skewness leads to a substantial generalization degradation (0.8391), suggesting that this feature is critical for cross-dataset transfer. The omission of normalization severely damages both metrics, underscoring its necessity for stable optimization.

\begin{table*}[htbp]
  \centering
  \caption{Ablation on Individual Features}
  \label{tab:feature_performance}
  \begin{tabular}{lccc}
    \toprule
    \textbf{Feature Selection} & \textbf{Acc.}& \textbf{Generalization Acc.} \\
    \midrule
    Only skewness     & 0.9605 & 0.8267  \\
    Only rotation\_ratio     & 0.9750 & 0.8463  \\
    Only skewness\_0     & 0.9760 & 0.8671 \\
    \hdashline
    w/o skewness & 0.9740 & 0.8391 \\
    w/o rotation\_ratio & 0.9810 & 0.8796 \\
    w/o skewness\_0 & 0.9660 & 0.8555\\
    \hdashline
    w/o normalize &0.9485 & 0.8250 \\
    \hdashline
    All Features & \textbf{0.9810} & \textbf{0.8914} \\
    \bottomrule
  \end{tabular}
\end{table*}

\paragraph{Ablation on SVD rank and classifier architecture}
In this ablation study, we aim to investigate the sensitivity of our binary classification task to the choice of the truncated rank r. Specifically, we apply truncated SVD to the weight matrices and vary the rank \( r \in \{8,16,32, 64, 128, 512\} \). The rotation\_ratio is fed into the same linear classifier used in the main experiments, keeping all other settings fixed to isolate the effect of rank. We evaluate in-domain accuracy on HaluEval and generalization accuracy across the four test sets. The resulting performance curves are plotted in Figure ~\ref{fig:Acc_over_SVD_Truncated_Ranks} of Appendix ~\ref{sec:More_Ablation_Results}. The in-domain accuracy (Acc) stays nearly constant (0.968–0.971) across all ranks, indicating that the task-relevant signal is highly concentrated in the top singular directions. In contrast, the generalization accuracy (Gen\_acc) varies more noticeably: it reaches its highest value of 0.8220 at r=128, while lower or higher ranks yield worse cross-domain performance. Importantly, the computational cost of truncated SVD grows with the rank. We measured the decomposition time for each setting: r=128 takes 649 minutes, whereas r=256 requires 1037 minutes, a nearly 60\% increase without improving generalization. Therefore, we select r=128 as the default rank, balancing optimal generalization and manageable computational overhead.

To investigate whether hallucination detection benefits from higher-capacity classifiers, we sweep the depth and width of the downstream MLP. Using the full feature set and the SVD rank of 128, we instantiate classifiers with \( L \) hidden layers, \( L \in \{2, 4, 8, 16\} \), and hidden dimension \( d_h \in \{512, 1024, 2048, 4096,8192\} \). All models are trained with identical hyperparameters, shown in Appendix \ref{sec:Ablation_on_classifier_architecture}, and we report in-domain and generalization accuracy. The complete results are shown in Fig ~\ref{fig:architec} of Appendix \ref{sec:More_Ablation_Results}.
We observe that within a certain range, increasing either the network depth or the width consistently yields higher performance. Notably, the performance gains demonstrate a clear trend of diminishing returns. The in-domain accuracy curves begin to saturate once the hidden size exceeds $4096$. A similar plateauing effect is observed in the Generalization Accuracy, where the margin of improvement between the 16-layer and 32-layer configurations becomes marginal across larger hidden dimensions (e.g., $d_h \ge 4096$). Interestingly, for the 32-layer variant, a slight degradation in-domain accuracy occurs when the hidden size reaches $8192$, potentially due to overfitting or optimization challenges associated with excessive depth and width. Balancing the trade-off between computational efficiency and classification performance, we select the detector configuration with $16$ layers and a hidden size of $4096$ as the default architecture for all main experiments and generalization studies.

\subsubsection{Which Layers Best Encode Hallucinations?}
To investigate which parts of the model most effectively capture hallucination-related signals, we divide the transformer layers into early, last, and all layers groups and examine several key modules, including the MLP, the attention block as a whole, and the individual projection inside attention and MLP. For each configuration, we extract the corresponding features and train a hallucination detector with our method under the same training setup. We then compare their hallucination detection performance on the same test sets. Results of LLaMA3-8B are shown in Tab \ref{tab:Ablation_on_Modules_of_LLaMA3-8B}, and those of Qwen3-8B are shown in Tab \ref{tab:Ablation_on_Modules_of_Qwen3-8B} of Appendix \ref{sec:More_Experimental_Results}.

\begin{table*}[htbp]
  \centering
  \caption{Ablation on Moudules of LLaMA3-8B}
  \label{tab:Ablation_on_Modules_of_LLaMA3-8B}
  \begin{tabular}{lccc}
    \toprule
    \textbf{Module Selection} & \textbf{Acc.}& \textbf{Generalization Acc.} \\
    \midrule
    Last Blocks + MLP & 0.9555 & 0.7645 \\
    Last Blocks + Attention & 0.9645 & 0.7941 \\
    Last Blocks + Both Modules & \textbf{0.9680} & \textbf{0.801} \\
    \hdashline
    Early Blocks + MLP & 0.8900 & 0.6730 \\
    Last Blocks + MLP & 0.9555 & 0.7653 \\
    All Blocks + MLP & \textbf{0.9630} & \textbf{0.7720} \\
    \hdashline
    Last Blocks + up\_proj & 0.9535 & 0.7235 \\
    Last Blocks + gate\_proj & 0.9450 & 0.7414 \\
    Last Blocks + down\_proj & 0.9440 & 0.7144 \\
    Last Blocks + q\_proj & 0.9375 & 0.6749 \\
    Last Blocks + k\_proj & 0.9325 & 0.6932 \\
    Last Blocks + v\_proj & \textbf{0.9615} & \textbf{0.7799} \\
    Last Blocks + o\_proj & 0.9515 & 0.7519 \\
    \hdashline
    Both Modules + All Blocks & \textbf{0.9760} & \textbf{0.8163} \\
    \bottomrule
  \end{tabular}
\end{table*}

\subsubsection{More Than QA}
To examine whether \MethodName{} generalizes beyond standard text QA-style hallucination detection, we further evaluate it on mathematical reasoning and vision-language hallucination scenarios. Specifically, we use MATH-Reasoning-Paths~\cite{zhou2026theoretical} to test hallucination detection in reasoning-intensive mathematical problems, and Hal-Eval~\cite{10.1145/3664647.3680576} to evaluate hallucinations in vision-language models. As shown in Table~\ref{tab:Out-Of-Distribution}, \MethodName{} maintains strong out-of-domain performance across both settings. These results suggest that the proposed weight-update features capture a general model-sample inconsistency signal rather than dataset-specific QA patterns.

\begin{table*}[t]
  \centering
  \caption{Generalization Performance across Out-Of-Distribution Datasets.}
  \label{tab:Out-Of-Distribution}
  \begin{adjustbox}{max width=\textwidth}
    \begin{tabular}{lccccc}
      \toprule
      \multirow{2}{*}{Model} 
      & \multirow{2}{*}{MATH-Reasoning-Paths}
      & \multicolumn{4}{c}{\textbf{\textit{Hal-Eval}}} \\ 
      \cline{3-6}
      &  & Objective & Attributive & Spatial-Relationship & Event \\
      \midrule
      \rowcolor{sealrow}
      \multicolumn{6}{l}{\textbf{\textit{Llama-3.2-11B-Vision-Instruct}}} \\
      Ours & 0.793 & 0.844 & 0.814 & 0.828  & 0.979  \\
      
      \midrule
      \rowcolor{sealrow}
      \multicolumn{6}{l}{\textbf{\textit{Qwen3.5-9B (it)}}} \\
      Ours(lately blocks) & 0.766 & 0.791  & 0.768 & 0.680 & 0.979 \\
      Ours(all blocks) &  0.766  & 0.79 & 0.795 & 0.704 & 0.985 \\
      \bottomrule
    \end{tabular}
  \end{adjustbox}
\end{table*}

\subsubsection{Generality across Model Scales and Architectures}
The previous ablation focuses on a single model, leaving open the question of whether the identified hallucination-sensitive signals and our detection method generalize across model sizes. In practical deployment, hallucination detectors may need to operate on LLMs ranging from lightweight models to large-scale models. To this end, we evaluate \MethodName{} under a unified protocol across models of different scales. For each tested model, we extract the same type of weight-update features from correct and hallucinated question-answer pairs and train a classifier using the identical procedure as in our main experiments. This setup isolates the effect of model scale while holding the data and training protocol fixed.

We select a diverse suite of dense LLMs spanning from 0.8B to 70B parameters, including Qwen3.5-0.8B-Base, Qwen3.5-2B-Base, Qwen3.5-4B-Base~\cite{qwen3.5}, Qwen3.5-9B-Base, Qwen3-14B-Base, Llama-3.2-1B-Instruct, Llama-3.2-3B-Instruct, Meta-Llama-3-8B-Instruct, and Llama-3.3-70B-Instruct. All models are evaluated on the same held-out test set. This comparison allows us to assess whether the proposed weight-update features remain effective as model capacity changes.
\begin{table*}[htbp]
  \centering
  \caption{Across Model Scales}
  \label{tab:Ablation_on_Moudules}
  \begin{tabular}{lcc}
    \toprule
    \textbf{Models} & \textbf{Acc.}& \textbf{Generalization Acc.} \\
    \midrule
    Qwen3.5-0.8B-Base & 0.9570 & 0.7683 \\
    Qwen3.5-2B-Base & 0.9755 & 0.8123 \\
    Qwen3.5-4B-Base & 0.9730 & 0.8464 \\
    Qwen3.5-9B-Base & 0.9575 & 0.8191 \\
    Qwen3-14B-Base & 0.9740 & 0.8155 \\
    \hdashline
    Llama-3.2-1B-Instruct & 0.9580 & 0.7728 \\
    Llama-3.2-3B-Instruct & 0.9605 & 0.7660 \\
    Meta-Llama-3-8B-Instruct & 0.9675  & 0.8026  \\
    Llama-3.3-70B-Instruct & 0.9690  & 0.8303  \\
    \bottomrule
  \end{tabular}
\end{table*}

\section{Conclusion}\label{sec:Conclusion}
In this paper, we proposed \MethodName{}, a hallucination detection framework that shifts the detection signal from static forward-pass representations to weight-update dynamics. By analyzing how sample-induced gradients interact with model weights, \MethodName{} captures two complementary features: the skewness of weight-gradient alignment and the SVD-based rotation ratio of the singular-vector basis. Extensive experiments across four model families and four benchmark datasets show that \MethodName{} achieves competitive in-domain performance and stronger cross-dataset generalization than existing white-box detectors. Further analyses demonstrate that the proposed features scale across model sizes from 0.8B to 70B parameters and transfer to out-of-domain settings, including mathematical reasoning and vision-language hallucination detection. These results suggest that weight-update dynamics provide a general and transferable signal for detecting hallucinations, filtering abnormal generations, and understanding how LLMs internally respond to faithful versus hallucinated content.
\iftaclpubformat

\section{Acknowledgments}

\else
\fi

\bibliography{References}

\begin{thebibliography}{20}
\expandafter\ifx\csname natexlab\endcsname\relax\def\natexlab#1{#1}\fi

\bibitem[{Chen et~al.(2024)Chen, Liu, Chen, Gu, Wu, Tao, Fu, and Ye}]{chen2024inside}
Chao Chen, Kai Liu, Ze~Chen, Yi~Gu, Yue Wu, Mingyuan Tao, Zhihang Fu, and Jieping Ye. 2024.
\newblock Inside: Llms' internal states retain the power of hallucination detection.
\newblock \emph{arXiv preprint arXiv:2402.03744}.

\bibitem[{Fadeeva et~al.(2024)Fadeeva, Rubashevskii, Shelmanov, Petrakov, Li, Mubarak, Tsymbalov, Kuzmin, Panchenko, Baldwin, Nakov, and Panov}]{fadeeva-etal-2024-fact}
Ekaterina Fadeeva, Aleksandr Rubashevskii, Artem Shelmanov, Sergey Petrakov, Haonan Li, Hamdy Mubarak, Evgenii Tsymbalov, Gleb Kuzmin, Alexander Panchenko, Timothy Baldwin, Preslav Nakov, and Maxim Panov. 2024.
\newblock \href {https://doi.org/10.18653/v1/2024.findings-acl.558} {Fact-checking the output of large language models via token-level uncertainty quantification}.
\newblock In \emph{Findings of the Association for Computational Linguistics: ACL 2024}, pages 9367--9385, Bangkok, Thailand. Association for Computational Linguistics.

\bibitem[{Farquhar et~al.(2024)Farquhar, Kossen, Kuhn, and Gal}]{farquhar2024detecting}
Sebastian Farquhar, Jannik Kossen, Lorenz Kuhn, and Yarin Gal. 2024.
\newblock Detecting hallucinations in large language models using semantic entropy.
\newblock \emph{Nature}, 630(8017):625--630.

\bibitem[{Grattafiori et~al.(2024)Grattafiori, Dubey, Jauhri, Pandey, Kadian, Al-Dahle, Letman, Mathur, Schelten, Vaughan et~al.}]{grattafiori2024llama}
Aaron Grattafiori, Abhimanyu Dubey, Abhinav Jauhri, Abhinav Pandey, Abhishek Kadian, Ahmad Al-Dahle, Aiesha Letman, Akhil Mathur, Alan Schelten, Alex Vaughan, et~al. 2024.
\newblock The llama 3 herd of models.
\newblock \emph{arXiv preprint arXiv:2407.21783}.

\bibitem[{Hu et~al.(2022)Hu, Shen, Wallis, Allen-Zhu, Li, Wang, Wang, and Chen}]{hu2022lora}
Edward~J Hu, Yelong Shen, Phillip Wallis, Zeyuan Allen-Zhu, Yuanzhi Li, Shean Wang, Lu~Wang, and Weizhu Chen. 2022.
\newblock \href {https://openreview.net/forum?id=nZeVKeeFYf9} {Lo{RA}: Low-rank adaptation of large language models}.
\newblock In \emph{International Conference on Learning Representations}.

\bibitem[{Hu et~al.(2025)Hu, Tu, Cheng, Li, Wang, Chen, Zhou, and Shan}]{hu2025harphallucinationdetectionreasoning}
Junjie Hu, Gang Tu, ShengYu Cheng, Jinxin Li, Jinting Wang, Rui Chen, Zhilong Zhou, and Dongbo Shan. 2025.
\newblock \href {http://arxiv.org/abs/2509.11536} {Harp: Hallucination detection via reasoning subspace projection}.

\bibitem[{Jiang et~al.(2024)Jiang, Jia, Dong, Ye, Xu, Yan, Zhang, and Zhang}]{10.1145/3664647.3680576}
Chaoya Jiang, Hongrui Jia, Mengfan Dong, Wei Ye, Haiyang Xu, Ming Yan, Ji~Zhang, and Shikun Zhang. 2024.
\newblock \href {https://doi.org/10.1145/3664647.3680576} {Hal-eval: A universal and fine-grained hallucination evaluation framework for large vision language models}.
\newblock In \emph{Proceedings of the 32nd ACM International Conference on Multimedia}, MM '24, page 525–534, New York, NY, USA. Association for Computing Machinery.

\bibitem[{Joshi et~al.(2017)Joshi, Choi, Weld, and Zettlemoyer}]{JoshiTriviaQA2017}
Mandar Joshi, Eunsol Choi, Daniel~S. Weld, and Luke Zettlemoyer. 2017.
\newblock Triviaqa: A large scale distantly supervised challenge dataset for reading comprehension.
\newblock In \emph{Proceedings of the 55th Annual Meeting of the Association for Computational Linguistics}, Vancouver, Canada. Association for Computational Linguistics.

\bibitem[{Li et~al.(2023)Li, Cheng, Zhao, Nie, and Wen}]{li-etal-2023-halueval}
Junyi Li, Xiaoxue Cheng, Xin Zhao, Jian-Yun Nie, and Ji-Rong Wen. 2023.
\newblock \href {https://doi.org/10.18653/v1/2023.emnlp-main.397} {{H}alu{E}val: A large-scale hallucination evaluation benchmark for large language models}.
\newblock In \emph{Proceedings of the 2023 Conference on Empirical Methods in Natural Language Processing}, pages 6449--6464, Singapore. Association for Computational Linguistics.

\bibitem[{Liu et~al.(2026)Liu, Khandelwal, Subramanian, Jouault, Rastogi, Sadé, Jeffares, Jiang, Cahill, Gavaudan, Sablayrolles, Héliou, You, Ehrenberg, Lo, Eliseev, Calvi, Sooriyarachchi, Bout, Rozière, Monicault, Lanfranchi, Barreau, Courtot, Grattarola, Dabert, de~las Casas, Chane-Sane, Ahmed, Berrada, Ecrepont, Guinet, Novikov, Kunsch, Lample, Martin, Gupta, Ludziejewski, Rute, Studnia, Amar, Delas, Roberts, Yadav, Chandu, Jain, Aitchison, Fainsin, Blier, Zhao, Martin, Saulnier, Gao, Buyl, Jennings, Pellat, Prins, Poirée, Guillaumin, Dinot, Futeral, Darrin, Augustin, Chiquier, Schimpf, Grinsztajn, Gupta, Raghuraman, Bousquet, Duchenne, Wang, von Platen, Jacob, Wambergue, Kurylowicz, Muddireddy, Chagniot, Stock, Agrawal, Torroba, Sauvestre, Soletskyi, Menneer, Vaze, Barry, Gandhi, Waghjale, Gandhi, Ghosh, Mishra, Aithal, Antoniak, Scao, Cachet, Sorg, Lavril, Saada, Chabal, Foubert, Robert, Wang, Lawson, Bewley, Bewley, Edwards, Jamil, Tomasini, Nemychnikova, Phung, Maladière, Richard, Bouaziz, Li,
  Marshall, Li, Yang, Ouahidi, Wang, Tang, and Ramzi}]{liu2026ministral3}
Alexander~H. Liu, Kartik Khandelwal, Sandeep Subramanian, Victor Jouault, Abhinav Rastogi, Adrien Sadé, Alan Jeffares, Albert Jiang, Alexandre Cahill, Alexandre Gavaudan, Alexandre Sablayrolles, Amélie Héliou, Amos You, Andy Ehrenberg, Andy Lo, Anton Eliseev, Antonia Calvi, Avinash Sooriyarachchi, Baptiste Bout, Baptiste Rozière, Baudouin~De Monicault, Clémence Lanfranchi, Corentin Barreau, Cyprien Courtot, Daniele Grattarola, Darius Dabert, Diego de~las Casas, Elliot Chane-Sane, Faruk Ahmed, Gabrielle Berrada, Gaëtan Ecrepont, Gauthier Guinet, Georgii Novikov, Guillaume Kunsch, Guillaume Lample, Guillaume Martin, Gunshi Gupta, Jan Ludziejewski, Jason Rute, Joachim Studnia, Jonas Amar, Joséphine Delas, Josselin~Somerville Roberts, Karmesh Yadav, Khyathi Chandu, Kush Jain, Laurence Aitchison, Laurent Fainsin, Léonard Blier, Lingxiao Zhao, Louis Martin, Lucile Saulnier, Luyu Gao, Maarten Buyl, Margaret Jennings, Marie Pellat, Mark Prins, Mathieu Poirée, Mathilde Guillaumin, Matthieu Dinot, Matthieu
  Futeral, Maxime Darrin, Maximilian Augustin, Mia Chiquier, Michel Schimpf, Nathan Grinsztajn, Neha Gupta, Nikhil Raghuraman, Olivier Bousquet, Olivier Duchenne, Patricia Wang, Patrick von Platen, Paul Jacob, Paul Wambergue, Paula Kurylowicz, Pavankumar~Reddy Muddireddy, Philomène Chagniot, Pierre Stock, Pravesh Agrawal, Quentin Torroba, Romain Sauvestre, Roman Soletskyi, Rupert Menneer, Sagar Vaze, Samuel Barry, Sanchit Gandhi, Siddhant Waghjale, Siddharth Gandhi, Soham Ghosh, Srijan Mishra, Sumukh Aithal, Szymon Antoniak, Teven~Le Scao, Théo Cachet, Theo~Simon Sorg, Thibaut Lavril, Thiziri~Nait Saada, Thomas Chabal, Thomas Foubert, Thomas Robert, Thomas Wang, Tim Lawson, Tom Bewley, Tom Bewley, Tom Edwards, Umar Jamil, Umberto Tomasini, Valeriia Nemychnikova, Van Phung, Vincent Maladière, Virgile Richard, Wassim Bouaziz, Wen-Ding Li, William Marshall, Xinghui Li, Xinyu Yang, Yassine~El Ouahidi, Yihan Wang, Yunhao Tang, and Zaccharie Ramzi. 2026.
\newblock \href {http://arxiv.org/abs/2601.08584} {Ministral 3}.

\bibitem[{Manakul et~al.(2023)Manakul, Liusie, and Gales}]{manakul-etal-2023-selfcheckgpt}
Potsawee Manakul, Adian Liusie, and Mark Gales. 2023.
\newblock \href {https://doi.org/10.18653/v1/2023.emnlp-main.557} {{S}elf{C}heck{GPT}: Zero-resource black-box hallucination detection for generative large language models}.
\newblock In \emph{Proceedings of the 2023 Conference on Empirical Methods in Natural Language Processing}, pages 9004--9017, Singapore. Association for Computational Linguistics.

\bibitem[{{Qwen Team}(2026)}]{qwen3.5}
{Qwen Team}. 2026.
\newblock \href {https://qwen.ai/blog?id=qwen3.5} {{Qwen3.5}: Towards native multimodal agents}.

\bibitem[{Rajpurkar et~al.(2018)Rajpurkar, Jia, and Liang}]{rajpurkar-etal-2018-know}
Pranav Rajpurkar, Robin Jia, and Percy Liang. 2018.
\newblock \href {https://doi.org/10.18653/v1/P18-2124} {Know what you don{'}t know: Unanswerable questions for {SQ}u{AD}}.
\newblock In \emph{Proceedings of the 56th Annual Meeting of the Association for Computational Linguistics (Volume 2: Short Papers)}, pages 784--789, Melbourne, Australia. Association for Computational Linguistics.

\bibitem[{Su et~al.(2024)Su, Wang, Ai, Hu, Wu, Zhou, and Liu}]{su-etal-2024-unsupervised}
Weihang Su, Changyue Wang, Qingyao Ai, Yiran Hu, Zhijing Wu, Yujia Zhou, and Yiqun Liu. 2024.
\newblock \href {https://doi.org/10.18653/v1/2024.findings-acl.854} {Unsupervised real-time hallucination detection based on the internal states of large language models}.
\newblock In \emph{Findings of the Association for Computational Linguistics: ACL 2024}, pages 14379--14391, Bangkok, Thailand. Association for Computational Linguistics.

\bibitem[{Team et~al.(2025)Team, Kamath, Ferret, Pathak, Vieillard, Merhej, Perrin, Matejovicova, Ramé, Rivière, Rouillard, Mesnard, Cideron, bastien Grill, Ramos, Yvinec, Casbon, Pot, Penchev, Liu, Visin, Kenealy, Beyer, Zhai, Tsitsulin, Busa-Fekete, Feng, Sachdeva, Coleman, Gao, Mustafa, Barr, Parisotto, Tian, Eyal, Cherry, Peter, Sinopalnikov, Bhupatiraju, Agarwal, Kazemi, Malkin, Kumar, Vilar, Brusilovsky, Luo, Steiner, Friesen, Sharma, Sharma, Gilady, Goedeckemeyer, Saade, Feng, Kolesnikov, Bendebury, Abdagic, Vadi, György, Pinto, Das, Bapna, Miech, Yang, Paterson, Shenoy, Chakrabarti, Piot, Wu, Shahriari, Petrini, Chen, Lan, Choquette-Choo, Carey, Brick, Deutsch, Eisenbud, Cattle, Cheng, Paparas, Sreepathihalli, Reid, Tran, Zelle, Noland, Huizenga, Kharitonov, Liu, Amirkhanyan, Cameron, Hashemi, Klimczak-Plucińska, Singh, Mehta, Lehri, Hazimeh, Ballantyne, Szpektor, Nardini, Pouget-Abadie, Chan, Stanton, Wieting, Lai, Orbay, Fernandez, Newlan, yeong Ji, Singh, Black, Yu, Hui, Vodrahalli, Greff, Qiu,
  Valentine, Coelho, Ritter, Hoffman, Watson, Chaturvedi, Moynihan, Ma, Babar, Noy, Byrd, Roy, Momchev, Chauhan, Sachdeva, Bunyan, Botarda, Caron, Rubenstein, Culliton, Schmid, Sessa, Xu, Stanczyk, Tafti, Shivanna, Wu, Pan, Rokni, Willoughby, Vallu, Mullins, Jerome, Smoot, Girgin, Iqbal, Reddy, Sheth, Põder, Bhatnagar, Panyam, Eiger, Zhang, Liu, Yacovone, Liechty, Kalra, Evci, Misra, Roseberry, Feinberg, Kolesnikov, Han, Kwon, Chen, Chow, Zhu, Wei, Egyed, Cotruta, Giang, Kirk, Rao, Black, Babar, Lo, Moreira, Martins, Sanseviero, Gonzalez, Gleicher, Warkentin, Mirrokni, Senter, Collins, Barral, Ghahramani, Hadsell, Matias, Sculley, Petrov, Fiedel, Shazeer, Vinyals, Dean, Hassabis, Kavukcuoglu, Farabet, Buchatskaya, Alayrac, Anil, Dmitry, Lepikhin, Borgeaud, Bachem, Joulin, Andreev, Hardin, Dadashi, and Hussenot}]{gemmateam2025gemma3technicalreport}
Gemma Team, Aishwarya Kamath, Johan Ferret, Shreya Pathak, Nino Vieillard, Ramona Merhej, Sarah Perrin, Tatiana Matejovicova, Alexandre Ramé, Morgane Rivière, Louis Rouillard, Thomas Mesnard, Geoffrey Cideron, Jean bastien Grill, Sabela Ramos, Edouard Yvinec, Michelle Casbon, Etienne Pot, Ivo Penchev, Gaël Liu, Francesco Visin, Kathleen Kenealy, Lucas Beyer, Xiaohai Zhai, Anton Tsitsulin, Robert Busa-Fekete, Alex Feng, Noveen Sachdeva, Benjamin Coleman, Yi~Gao, Basil Mustafa, Iain Barr, Emilio Parisotto, David Tian, Matan Eyal, Colin Cherry, Jan-Thorsten Peter, Danila Sinopalnikov, Surya Bhupatiraju, Rishabh Agarwal, Mehran Kazemi, Dan Malkin, Ravin Kumar, David Vilar, Idan Brusilovsky, Jiaming Luo, Andreas Steiner, Abe Friesen, Abhanshu Sharma, Abheesht Sharma, Adi~Mayrav Gilady, Adrian Goedeckemeyer, Alaa Saade, Alex Feng, Alexander Kolesnikov, Alexei Bendebury, Alvin Abdagic, Amit Vadi, András György, André~Susano Pinto, Anil Das, Ankur Bapna, Antoine Miech, Antoine Yang, Antonia Paterson, Ashish
  Shenoy, Ayan Chakrabarti, Bilal Piot, Bo~Wu, Bobak Shahriari, Bryce Petrini, Charlie Chen, Charline~Le Lan, Christopher~A. Choquette-Choo, CJ~Carey, Cormac Brick, Daniel Deutsch, Danielle Eisenbud, Dee Cattle, Derek Cheng, Dimitris Paparas, Divyashree~Shivakumar Sreepathihalli, Doug Reid, Dustin Tran, Dustin Zelle, Eric Noland, Erwin Huizenga, Eugene Kharitonov, Frederick Liu, Gagik Amirkhanyan, Glenn Cameron, Hadi Hashemi, Hanna Klimczak-Plucińska, Harman Singh, Harsh Mehta, Harshal~Tushar Lehri, Hussein Hazimeh, Ian Ballantyne, Idan Szpektor, Ivan Nardini, Jean Pouget-Abadie, Jetha Chan, Joe Stanton, John Wieting, Jonathan Lai, Jordi Orbay, Joseph Fernandez, Josh Newlan, Ju~yeong Ji, Jyotinder Singh, Kat Black, Kathy Yu, Kevin Hui, Kiran Vodrahalli, Klaus Greff, Linhai Qiu, Marcella Valentine, Marina Coelho, Marvin Ritter, Matt Hoffman, Matthew Watson, Mayank Chaturvedi, Michael Moynihan, Min Ma, Nabila Babar, Natasha Noy, Nathan Byrd, Nick Roy, Nikola Momchev, Nilay Chauhan, Noveen Sachdeva, Oskar
  Bunyan, Pankil Botarda, Paul Caron, Paul~Kishan Rubenstein, Phil Culliton, Philipp Schmid, Pier~Giuseppe Sessa, Pingmei Xu, Piotr Stanczyk, Pouya Tafti, Rakesh Shivanna, Renjie Wu, Renke Pan, Reza Rokni, Rob Willoughby, Rohith Vallu, Ryan Mullins, Sammy Jerome, Sara Smoot, Sertan Girgin, Shariq Iqbal, Shashir Reddy, Shruti Sheth, Siim Põder, Sijal Bhatnagar, Sindhu~Raghuram Panyam, Sivan Eiger, Susan Zhang, Tianqi Liu, Trevor Yacovone, Tyler Liechty, Uday Kalra, Utku Evci, Vedant Misra, Vincent Roseberry, Vlad Feinberg, Vlad Kolesnikov, Woohyun Han, Woosuk Kwon, Xi~Chen, Yinlam Chow, Yuvein Zhu, Zichuan Wei, Zoltan Egyed, Victor Cotruta, Minh Giang, Phoebe Kirk, Anand Rao, Kat Black, Nabila Babar, Jessica Lo, Erica Moreira, Luiz~Gustavo Martins, Omar Sanseviero, Lucas Gonzalez, Zach Gleicher, Tris Warkentin, Vahab Mirrokni, Evan Senter, Eli Collins, Joelle Barral, Zoubin Ghahramani, Raia Hadsell, Yossi Matias, D.~Sculley, Slav Petrov, Noah Fiedel, Noam Shazeer, Oriol Vinyals, Jeff Dean, Demis Hassabis,
  Koray Kavukcuoglu, Clement Farabet, Elena Buchatskaya, Jean-Baptiste Alayrac, Rohan Anil, Dmitry, Lepikhin, Sebastian Borgeaud, Olivier Bachem, Armand Joulin, Alek Andreev, Cassidy Hardin, Robert Dadashi, and Léonard Hussenot. 2025.
\newblock \href {http://arxiv.org/abs/2503.19786} {Gemma 3 technical report}.

\bibitem[{Yang et~al.(2025)Yang, Li, Yang, Zhang, Hui, Zheng, Yu, Gao, Huang, Lv, Zheng, Liu, Zhou, Huang, Hu, Ge, Wei, Lin, Tang, Yang, Tu, Zhang, Yang, Yang, Zhou, Zhou, Lin, Dang, Bao, Yang, Yu, Deng, Li, Xue, Li, Zhang, Wang, Zhu, Men, Gao, Liu, Luo, Li, Tang, Yin, Ren, Wang, Zhang, Ren, Fan, Su, Zhang, Zhang, Wan, Liu, Wang, Cui, Zhang, Zhou, and Qiu}]{yang2025qwen3technicalreport}
An~Yang, Anfeng Li, Baosong Yang, Beichen Zhang, Binyuan Hui, Bo~Zheng, Bowen Yu, Chang Gao, Chengen Huang, Chenxu Lv, Chujie Zheng, Dayiheng Liu, Fan Zhou, Fei Huang, Feng Hu, Hao Ge, Haoran Wei, Huan Lin, Jialong Tang, Jian Yang, Jianhong Tu, Jianwei Zhang, Jianxin Yang, Jiaxi Yang, Jing Zhou, Jingren Zhou, Junyang Lin, Kai Dang, Keqin Bao, Kexin Yang, Le~Yu, Lianghao Deng, Mei Li, Mingfeng Xue, Mingze Li, Pei Zhang, Peng Wang, Qin Zhu, Rui Men, Ruize Gao, Shixuan Liu, Shuang Luo, Tianhao Li, Tianyi Tang, Wenbiao Yin, Xingzhang Ren, Xinyu Wang, Xinyu Zhang, Xuancheng Ren, Yang Fan, Yang Su, Yichang Zhang, Yinger Zhang, Yu~Wan, Yuqiong Liu, Zekun Wang, Zeyu Cui, Zhenru Zhang, Zhipeng Zhou, and Zihan Qiu. 2025.
\newblock \href {http://arxiv.org/abs/2505.09388} {Qwen3 technical report}.

\bibitem[{Yang et~al.(2018)Yang, Qi, Zhang, Bengio, Cohen, Salakhutdinov, and Manning}]{yang2018hotpotqa}
Zhilin Yang, Peng Qi, Saizheng Zhang, Yoshua Bengio, William~W. Cohen, Ruslan Salakhutdinov, and Christopher~D. Manning. 2018.
\newblock {HotpotQA}: A dataset for diverse, explainable multi-hop question answering.
\newblock In \emph{Conference on Empirical Methods in Natural Language Processing ({EMNLP})}.

\bibitem[{Yehuda et~al.(2024)Yehuda, Malkiel, Barkan, Weill, Ronen, and Koenigstein}]{yehuda-etal-2024-interrogatellm}
Yakir Yehuda, Itzik Malkiel, Oren Barkan, Jonathan Weill, Royi Ronen, and Noam Koenigstein. 2024.
\newblock \href {https://doi.org/10.18653/v1/2024.acl-long.506} {{I}nterrogate{LLM}: Zero-resource hallucination detection in {LLM}-generated answers}.
\newblock In \emph{Proceedings of the 62nd Annual Meeting of the Association for Computational Linguistics (Volume 1: Long Papers)}, pages 9333--9347, Bangkok, Thailand. Association for Computational Linguistics.

\bibitem[{Zhang et~al.(2025)Zhang, Hu, Zhang, Zhang, and Wan}]{zhang-etal-2025-icr}
Zhenliang Zhang, Xinyu Hu, Huixuan Zhang, Junzhe Zhang, and Xiaojun Wan. 2025.
\newblock \href {https://doi.org/10.18653/v1/2025.acl-long.880} {{ICR} probe: Tracking hidden state dynamics for reliable hallucination detection in {LLM}s}.
\newblock In \emph{Proceedings of the 63rd Annual Meeting of the Association for Computational Linguistics (Volume 1: Long Papers)}, pages 17986--18002, Vienna, Austria. Association for Computational Linguistics.

\bibitem[{Zhou et~al.(2026)Zhou, Yuhao, Li, Yao, Guo, Li, and Ma}]{zhou2026theoretical}
Zhi Zhou, Tan Yuhao, Zenan Li, Yuan Yao, Lan-Zhe Guo, Yu-Feng Li, and Xiaoxing Ma. 2026.
\newblock A theoretical study on bridging internal probability and self-consistency for llm reasoning.
\newblock \emph{Advances in Neural Information Processing Systems}, 38:87380--87413.

\end{thebibliography}
\bibliographystyle{acl_natbib}

\onecolumn

\appendix

\section{More Experimental Details}

\subsection{Datasets}
For each dataset, we randomly sample 10,000 instances and split them into training and test sets with a ratio of 0.8 to 0.2. Specifically, we use the qa\_data subset of HaluEval and randomly sample instances labeled as either correct or incorrect responses, while ensuring balanced label distributions in both the training and test sets.  \MethodName{} and all baselines are evaluated on the same sampled data to ensure a fair comparison.

\subsection{Baselines}
\begin{itemize}
  \item MIND~\cite{su-etal-2024-unsupervised}:We treat the complete supporting context together with the user query as $w_i^{\prime}$ and the answer to the query as $e_i$. If the beginning of the first sentence $G_i$ generated by the LLM matches $e_i$, we regard the generation as non-hallucinatory. Conversely, if $G_i$ does not start with $e_i$, we consider it to exhibit hallucination.
  \item InterrogateLLM~\cite{yehuda-etal-2024-interrogatellm}:We evaluate InterrogateLLM in its single-model configuration, where the same LLM serves as both the forward generator and the backward reconstructor. This ensures a fair comparison with SKEW-SVD, as both methods operate under the same resource constraint of accessing only one model.
\end{itemize}

\subsection{All Experimental Setups}

\subsubsection{Main Results}\label{sec:Appendix_Main_results}

\subsubsection{Ablation on classifier architecture}\label{sec:Ablation_on_classifier_architecture}

\section{More Experimental Results}\label{sec:More_Experimental_Results}

\subsection{generalization\_appendix}\label{sec:generalization_appendix}

\subsection{More Ablation Results}\label{sec:More_Ablation_Results}

\begin{table*}[htbp]
  \centering
  \caption{Ablation on Modules of Qwen3-8B}
  \label{tab:Ablation_on_Modules_of_Qwen3-8B}
  \begin{tabular}{lccc}
    \toprule
    \textbf{Module Selection} & \textbf{Acc.}& \textbf{Generalization Acc.} \\
    \midrule
    Last Blocks + up\_proj & 0.9255 & 0.7740 \\
    Last Blocks + gate\_proj & 0.9335 & 0.7658 \\
    Last Blocks + down\_proj & 0.9170 & 0.7723 \\
    Last Blocks + q\_proj & 0.9215 & 0.7375 \\
    Last Blocks + k\_proj & 0.9050 & 0.7094 \\
    Last Blocks + v\_proj & \textbf{0.9440} & \textbf{0.7845} \\
    Last Blocks + o\_proj & 0.9325 & 0.7703 \\
    \hdashline
    All Blocks + Both Modules & \textbf{0.9530}  & \textbf{0.8388} \\
    \bottomrule
  \end{tabular}
\end{table*}

\begin{figure}[ht]
  \centering
  \includegraphics[width=\linewidth]{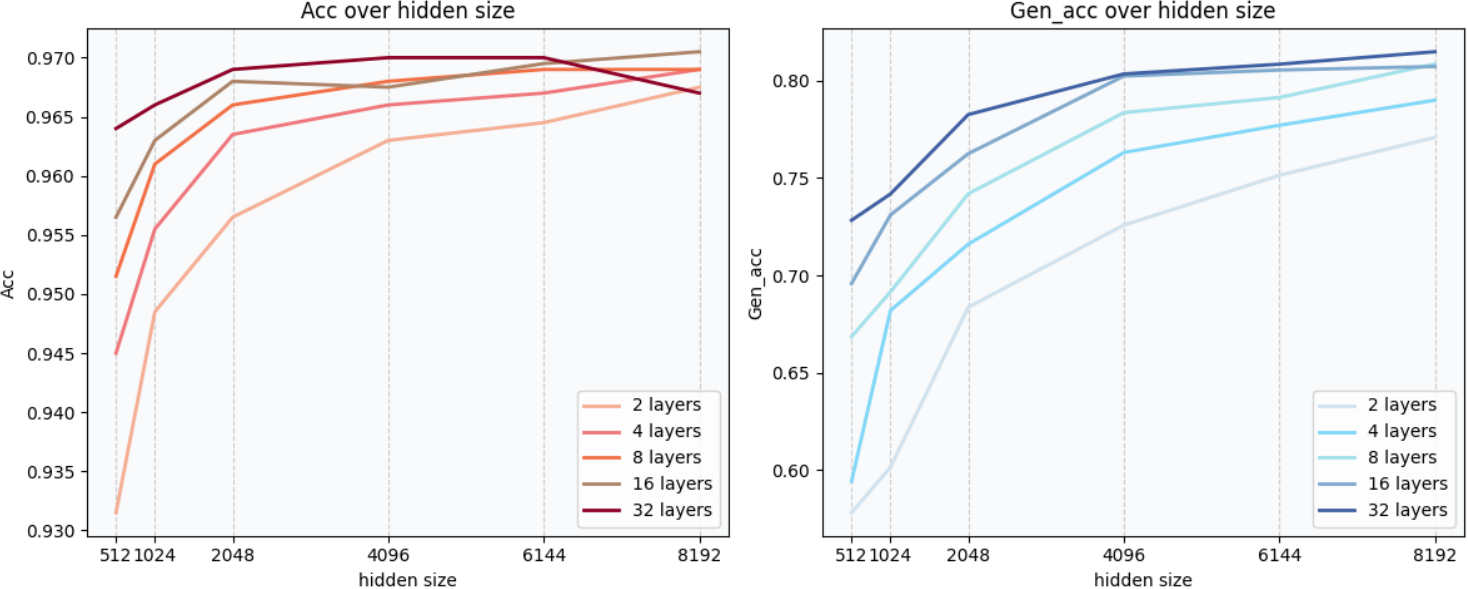}
  \caption{Ablation on classifier architecture}
  \label{fig:architec}
\end{figure}

\begin{figure}[ht]
  \centering
  \includegraphics[width=0.5\linewidth]{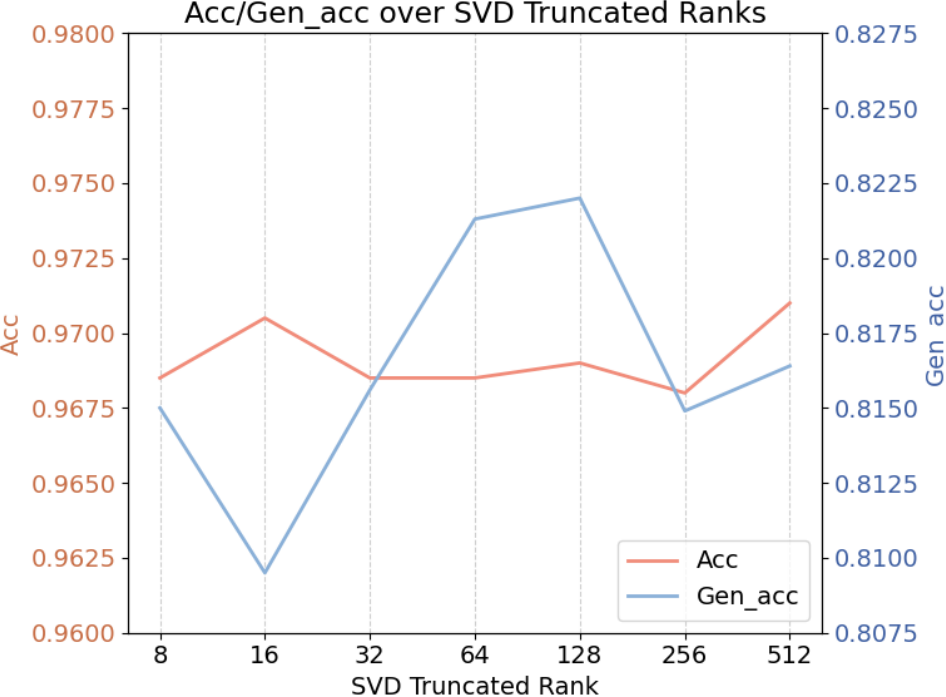}
  \caption{Acc/Generalization Acc over SVD Truncated Ranks}
  \label{fig:Acc_over_SVD_Truncated_Ranks}
\end{figure}

\end{document}